\documentclass{article}
\pdfoutput=1
\usepackage{ijcai20}

\usepackage{times}

\usepackage{soul}
\usepackage{url}
\usepackage{hyperref}
\usepackage[utf8]{inputenc}
\usepackage[small]{caption}
\usepackage{graphicx}
\usepackage{amsmath}
\usepackage{booktabs}
\usepackage{subfigure}
\usepackage{verbatim}
\usepackage{graphics}
\usepackage{multirow}
\usepackage{multicol}
\usepackage{natbib}
\newcommand{\bert} {{\tt BERT}}
\newcommand{\bertbase} {\tt BERT\textsubscript{BASE}}

\newcommand{\albert} {{\tt ALBERT}}

\newcommand{\pb} {{\tt PoWER-BERT}}
\newcommand{\distilbert} {{\tt DistilBERT}}
\newcommand{\bertpkd} {{\tt BERT-PKD}}
\newcommand{\headprune} {{\tt Head-Prune}}

\newcommand{\cls} {{\sf CLS}}
\newcommand{\pad} {{\tt PAD}}

\newcommand{\headws} {{\tt Head-WS}}
\newcommand{\randws} {{\tt Rand-WS}}
\newcommand{\attnws} {{\tt Attn-WS}}
\newcommand{\matM} {\mathbf{M}}
\newcommand{\matE} {\mathbf{E}}
\newcommand{\wqh} {\mathbf{W}^h_{\mathbf{q}}}
\newcommand{\wkh} {\mathbf{W}^h_{\mathbf{k}}}
\newcommand{\wvh} {\mathbf{W}^h_{\mathbf{v}}}

\newcommand{\matAh} {\mathbf{A}_h}
\newcommand{\matVh} {\mathbf{V}_h}
\newcommand{\matZh} {\mathbf{Z}_h}

\newcommand{\softmax} {\sf{softmax}}
\newcommand{\bfX} {\mathbf{X}}
\newcommand{\bfY} {\mathbf{Y}}
\newcommand{\entropy} {\mathbf{H}}
\newcommand{\MI} {\mathbf{MI}}
\newcommand{\sig} {{\tt Sig}}
\newcommand{\sigidx}{{\tt Sig}^{\it pos}}
\newcommand{\sigparam} {{\mathbf r}}
\newcommand{\mass} {{\rm mass}}

\newcommand{\calL} {{\cal L}}

\newcommand{\mytheta} {\mathbf{\Theta}}
\newcommand{\extract} {\tt extract}
\newcommand{\softextract} {\tt soft-extract}
\newcommand{\maxl} {N}

\usepackage{fancyhdr}
\title{PoWER-BERT: Accelerating BERT Inference via \\ Progressive Word-vector Elimination}

\author{
Saurabh Goyal$^1$\footnote{Contact Author}\and
Anamitra R. Choudhury$^1$\and
Saurabh M. Raje$^1$\and
Venkatesan T. Chakaravarthy$^1$\and
Yogish Sabharwal$^1$\And
Ashish Verma$^2$\\
\affiliations
$^1$IBM Research India\\
$^2$IBM Research USA\\
\emails
\{saurago1, anamchou, saraje01, vechakra, ysabharwal\}@in.ibm.com,
ashish.verma1@ibm.com
}

\begin{document}

\maketitle

\begin{abstract}
We develop a novel method, called {\pb}, for improving
the inference time of the popular {\bert} model, while maintaining the accuracy.
It works by:
a) exploiting redundancy pertaining to word-vectors (intermediate 
encoder outputs) and eliminating the redundant vectors.
b) determining which word-vectors to eliminate by developing a strategy for measuring their significance, based on the self-attention mechanism.
c) learning how many word-vectors to eliminate by augmenting the {\bert} model and the loss function.
Experiments on the standard GLUE benchmark shows that {\pb} achieves up to $4.5$x reduction in inference time over {\bert} with $<1\%$ loss in accuracy. We show that {\pb} offers significantly better trade-off between accuracy and inference time compared to prior methods.
We demonstrate that our method attains up to $6.8$x reduction in inference time with $<1\%$ loss in accuracy when applied over {\albert}, a highly compressed version of {\bert}. The code for {\pb} is publicly available at \url{https://github.com/IBM/PoWER-BERT}.
\end{abstract}
\section{Introduction}
The {\bert} model \cite{bert} has gained popularity as an effective approach for natural language processing. 
It has achieved significant success on standard benchmarks such as
GLUE \cite{glue} and SQuAD \cite{qnli},
dealing with sentiment classification, question-answering, natural language inference and language acceptability tasks.
The model has been used in applications ranging from text summarization \cite{summarization} to biomedical text mining \cite{biomine}.

\begin{figure*}[t]
\center
\includegraphics[width=7.0in]{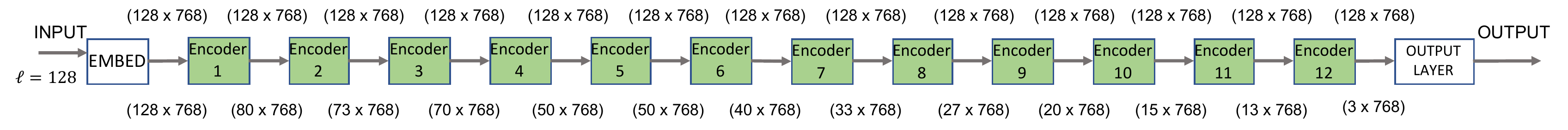}
\caption{
Illustration of {\pb} scheme over {\bertbase} that has $L=12$ encoders and hidden size $H=768$. The words are first embedded as vectors of length $H=768$. 
The numbers show output sizes for each encoder for input sequence of length $\maxl=128$. 
The numbers on the top and the bottom correspond to {\bertbase} and {\pb}, respectively.
In this example, the first encoder eliminates $48$ and retains $80$ word-vectors, whereas the second eliminates $7$ more and retains $73$ word-vectors. 
The hidden size remains at $768$.
}
\label{fig:snake}
\end{figure*}

The {\bert} model consists of an embedding layer, a chain of encoders and an output layer. 
The input words are first embedded as vectors, which are then transformed by the pipeline of encoders and the final prediction is derived at the output layer (see Figure \ref{fig:snake}). The model is known to be compute intensive, resulting in high infrastructure demands and latency, whereas low latency 
is vital for a good customer experience. Therefore, it is crucial to design methods that reduce the computational demands of {\bert} in order to successfully meet the latency 
and resource requirements of a production environment. 

Consequently, recent studies have focused on optimizing two fundamental metrics: 
model size and inference time.
The recently proposed {\albert} \cite{albert} achieves significant compression over {\bert}
by sharing parameters across the encoders and decomposing the embedding layer.
However, there is almost no impact on the inference time, 
since the amount of computation remains the same during inference
(even though training is faster).

Other studies have aimed for optimizing both the metrics simultaneously.
Here, a natural strategy is to reduce the number of encoders
and the idea has been employed by {\distilbert} \cite{distil-bert} and {\bertpkd} \cite{bert-pkd} 
within the knowledge distillation paradigm.
An alternative approach is to shrink the individual encoders.
Each encoder comprises of multiple self-attention heads and 
the {\headprune} strategy \cite{head-prune} removes a fraction of the heads 
by measuring their significance.
In order to achieve considerable reduction in the two metrics,
commensurate number of encoders/heads have to be pruned,
and the process leads to noticeable loss in accuracy.
The above approaches operate by removing the redundant model parameters
using strategies such as parameter sharing and encoder/attention-head removal.

\paragraph{Our Objective and Approach.}
We target the metric of inference time for a wide range of classification tasks.
The objective is to achieve significant reduction on the metric,
while maintaining the accuracy, and derive improved trade-off between the two.

In contrast to the prior approaches, we keep the model parameters intact.
Instead, we identify and exploit a different type of redundancy 
that pertains to the 
the intermediate vectors computed along the encoder pipeline,
which we henceforth denote as {\em word-vectors}.
We demonstrate that, due to the self-attention mechanism, 
there is diffusion of information: as the word-vectors pass 
through the encoder pipeline, they start carrying similar information, resulting in redundancy. Consequently, a significant fraction of the word-vectors 
can be eliminated in a progressive manner as we move from the 
first to the last encoder. 
The removal of the word-vectors reduces the computational load and results in improved inference time.
Based on the above ideas, we develop a novel scheme called 
{\pb} (\textbf{P}r\textbf{o}gressive \textbf{W}ord-vector \textbf{E}limination for inference time \textbf{R}eduction of \textbf{BERT}).
Figure \ref{fig:snake} presents an illustration.


\paragraph{Main Contributions.} Our main contributions are summarized below.
\begin{itemize}
\item
We develop a novel scheme called {\pb} for improving {\bert} inference time.
It is based on exploiting a new type of redundancy within the {\bert} model 
pertaining to the word-vectors. 
As part of the scheme, we design strategies for determining how many and which word-vectors to eliminate at each encoder.
\item
We present an experimental evaluation on a wide spectrum of classification/regression tasks from the popular GLUE benchmark.
The results show that {\pb} achieves up to $4.5$x 
reduction in inference time over {\bertbase} with $<1\%$ loss in accuracy.
\item 
We perform a comprehensive comparison with the state-of-the-art inference time reduction methods 
and demonstrate that {\pb} offers significantly better trade-off between inference time and accuracy. 
\item 
We show that our scheme can also be used to accelerate {\albert}, 
a highly compressed variant of {\bert},
yielding up to $6.8$x reduction in inference time. The code for {\pb} is publicly available at \url{https://github.com/IBM/PoWER-BERT}.
\end{itemize}

\paragraph{Related Work.}
In general, different methods for deep neural network compression have been developed such as pruning network connections \cite{prune,molchanov2017variational}, 
pruning filters/channels from the convolution layers \cite{he2017channel,molchanov2016pruning},
weight quantization \cite{quantize}, 
knowledge distillation from teacher to student model \cite{hinton2015distilling,sau2016deep}
and singular value decomposition of weight matrices \cite{svd,kim2015compression}.

Some of these general techniques have been explored for {\bert}:
weight quantization \cite{shen2019qbert,zafrir2019q8bert}, 
structured weight pruning \cite{wang2019structured} 
and dimensionality reduction \cite{albert,wang2019structured}. 
Although these techniques offer significant model size reduction,
they do not result in proportional inference time gains and 
some of them require specific hardware to execute. 
Another line of work has exploited pruning entries of the attention matrices 
\cite{sparse-transformer,Correia_2019,Peters_2019,martins2016softmax}.
However, the goal of these work is to improve translation accuracy,
they do not result in either model size or inference time reduction.
The {\bert} model allows for compression via other methods:
sharing of encoder parameters \cite{albert}, removing encoders via distillation \cite{distil-bert,bert-pkd,liu2019improving}, 
and pruning attention heads \cite{head-prune,mccarley2019pruning}.

Most of these prior approaches are based on removing redundant parameters.
{\pb} is an orthogonal technique that retains all the parameters, 
and eliminates only the redundant word-vectors.
Consequently, the scheme can be applied over and used to accelerate
inference of compressed models.
Our experimental evaluation demonstrates the phenomenon
by applying the scheme over {\albert}.

In terms of inference time, removing an encoder can be considered equivalent to eliminating all its output word-vectors. However, 
encoder elimination is a coarse-grained mechanism that removes the encoders in totality.
To achieve considerable gain on inference time, 
a commensurate number of encoders need to pruned, leading to accuracy loss. 
In contrast, word-vector elimination is a fine-grained method that keeps the encoders intact and eliminates only a fraction of word-vectors. 
Consequently, as demonstrated in our
experimental study, word-vector elimination leads to improved inference time gains.

\section{Background}
In this section, we present an overview of the {\bert} model focusing on the aspects that are essential to  our discussion.
Throughout the paper, we consider the {\bertbase} version with $L=12$ encoders, $A=12$ self-attention heads per encoder 
and hidden size $H=768$. The techniques can be readily applied to other versions.

The inputs in the dataset get tokenized and augmented with a {\cls} token at the beginning. 
A suitable maximum length $\maxl$ is chosen,
and shorter input sequences get padded to achieve an uniform length of $\maxl$.

Given an input of length $\maxl$, each word first gets embedded as a vector of length $H=768$. 
The word-vectors are then transformed by the chain of encoders
using a self-attention mechanism that captures information from the other word-vectors.
At the output layer, the final prediction is derived from the vector corresponding to the {\cls} token
and the other word-vectors are ignored.
{\pb} utilizes the self-attention mechanism to measure the significance of the word-vectors.
This mechanism is described below.

\paragraph{Self-Attention Mechanism.}
Each encoder comprises of a self-attention module consisting of $12$ attention heads and a feed-forward network. 
Each head $h\in [1,12]$ is associated
with three weight matrices $\wqh$, $\wkh$ and $\wvh$, called the query, the key and the value matrices. 

Let $\matM$ be the matrix of size $\maxl\times 768$ input to the encoder.
Each head $h$ computes an {\em attention matrix}:
\[
	\matAh = {\softmax}[(\matM\times \wqh)\times (\matM\times \wkh)^T]
\]
with {\softmax} applied row-wise. The attention matrix $\matAh$ is of size $\maxl\times \maxl$, wherein each row sums to $1$.
The head computes matrices $\matVh = \matM\times \wvh$ and $\matZh = \matAh\times \matVh$.
The encoder concatenates the $\matZh$ matrices over all the heads 
and derives its output after further processing.


\section{PoWER-BERT Scheme}
\subsection{Motivation}
\label{sec:motive}
{\bert} derives the final prediction from the word-vector corresponding to the {\cls} token.
We conducted experiments to determine whether it is critical to derive the final prediction from the {\cls} token during inference.
The results over different datasets showed that other word positions can be used as well, with minimal variations in accuracy.
For instance, on the SST-2 dataset from our experimental study,
the mean drop in accuracy across the different positions was only 
$1.2\%$ with a standard deviation of $0.23\%$ (compared to baseline accuracy of $92.43\%$).
We observed that the fundamental reason was diffusion of information.

\begin{figure}[t!]
\center
\resizebox{0.70\linewidth}{!}{
\includegraphics[width=1.7in]{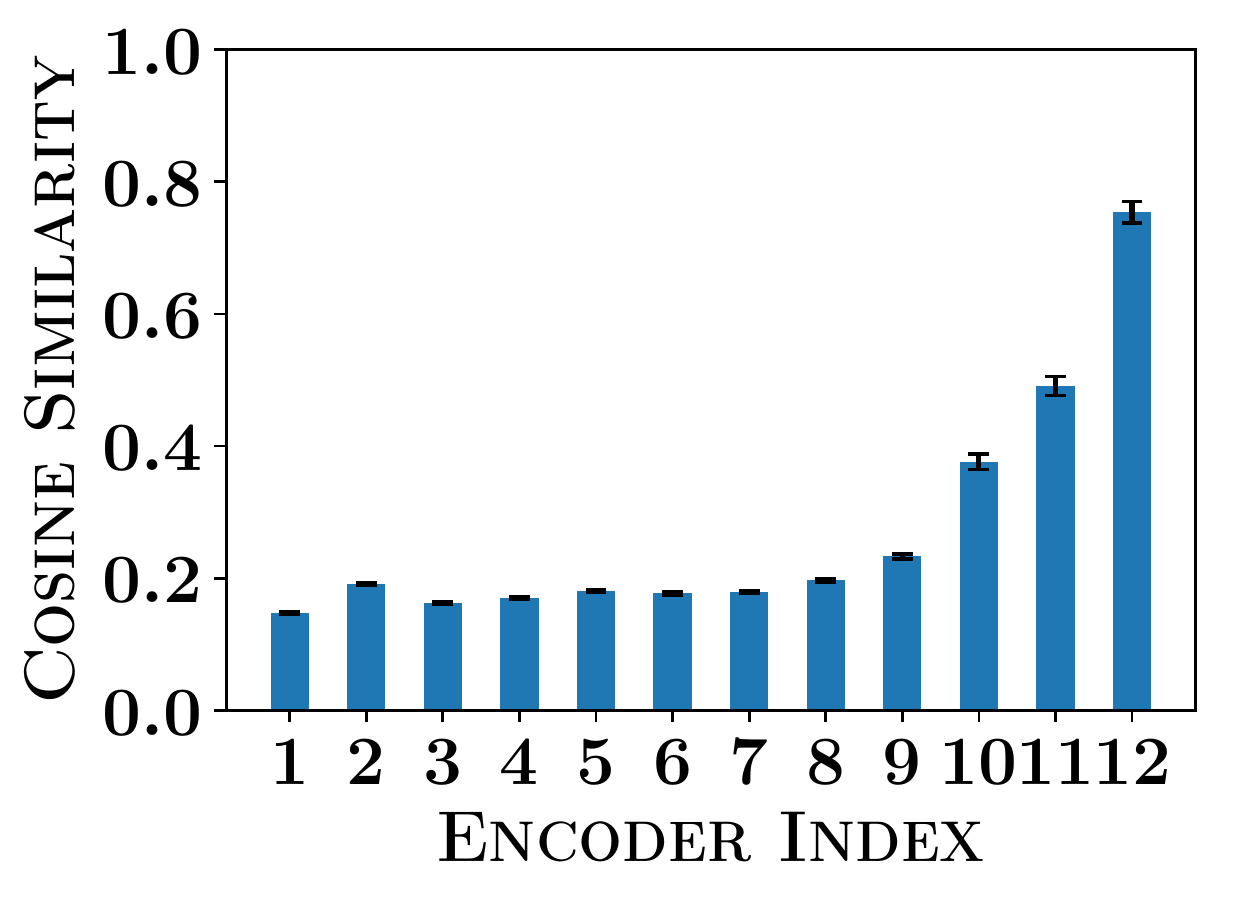}
}
\caption{Cosine similarity for {\bert} encoders on the SST-2 dataset. The $j^{th}$ bar represents cosine similarity for the $j^{th}$ encoder, averaged over all pairs of word-vectors and all inputs.}
\label{fig:cosine}
\end{figure}

\paragraph{Diffusion of Information.}
As the word-vectors pass through the encoder pipeline, 
they start progressively carrying similar information due to the self-attention mechanism.
We demonstrate the phenomenon through cosine similarity measurements.
Let $j\in [1,12]$ be an encoder. 
For each input, compute the cosine similarity between each of the $\binom{\maxl}{2}$ 
pairs of word-vectors output by the encoder, where $\maxl$ is the input length. 
Compute the average over all pairs and all inputs in the dataset.
As an illustration, Figure \ref{fig:cosine} shows the results for the SST-2 dataset.
We observe that the similarity increases with the  encoder index, implying diffusion of information. 
The diffusion leads to redundancy of the word-vectors and the model is able to 
derive the final prediction from any word-vector at the output layer.

The core intuition behind {\pb} 
is that the redundancy of the word-vectors cannot possibly manifest abruptly at the last layer, rather must build progressively through the encoder pipeline.
Consequently, we should be able to eliminate word-vectors in a progressive manner across all the encoders.

\paragraph{PoWER-BERT Components.} 
The {\pb} scheme involves two critical, inter-related tasks.
First, we identify a {\em retention configuration}:
a monotonically decreasing sequence $(\ell_1, \ell_2, \ldots, \ell_{12})$
that specifies the number of word-vectors $\ell_j$ to retain at encoder $j$.
For example, in Figure \ref{fig:snake}, the configuration is
$(80, 73, 70, 50, 50, 40, 33, 27, 20, 15, 13, 3)$. 
Secondly, we do {\em word-vector selection}, i.e., 
for a given input, determine which $\ell_j$ word-vectors to retain at each encoder $j$.
We first address the task of word-vector selection.

\subsection{Word-vector Selection}
\label{sec:wsel}
Assume that we are given a retention configuration $(\ell_1, \ell_2, \ldots, \ell_{12})$. 
Consider an encoder $j\in [1,12]$. The input to the encoder is a collection of $\ell_{j-1}$
word-vectors arranged in the form of a matrix of size $\ell_{j-1}\times 768$
(taking $\ell_0 = \maxl$). Our aim is to select $\ell_j$ word-vectors to retain
and we consider two kinds of strategies.

\paragraph{Static and Dynamic Strategies.}
Static strategies fix $\ell_j$ positions and retain the word-vectors at the same positions across all the input sequences in the dataset.
A natural static strategy is to retain the first (or head) $\ell_j$ word-vectors.
The intuition is that the input sequences are of varying lengths and an uniform length of $\maxl$ 
is achieved by adding {\pad} tokens that carry little information.
The strategy aims to remove as many {\pad} tokens on the average as possible,
even though actual word-vectors may also get eliminated.
A related method is to fix $\ell_j$ positions at random and retain word-vectors only at those positions across the dataset.
We denote these strategies as {\headws} and {\randws}, respectively (head/random word-vector selection).

In contrast to the static strategies, the dynamic strategies select the positions on a per-input basis.
While the word-vectors tend to carry similar information at the final encoders, 
in the earlier encoders, they have different levels of influence over the final prediction.
The positions of the significant word-vectors vary across the dataset.
Hence, it is a better idea to select the positions for each input independently,
as confirmed by our experimental evaluation.

We develop a scoring mechanism for estimating the significance of the word-vectors
satisfying the following criterion:
the score of a word-vector must be positively correlated with its influence on the final classification output
(namely, word-vectors of higher influence get higher score).
We accomplish the task by utilizing the self-attention mechanism
and design a dynamic strategy, denoted as {\attnws}.

\begin{figure}[t]
\center
\resizebox{0.85\linewidth}{!}{
\includegraphics[width=1.5in]{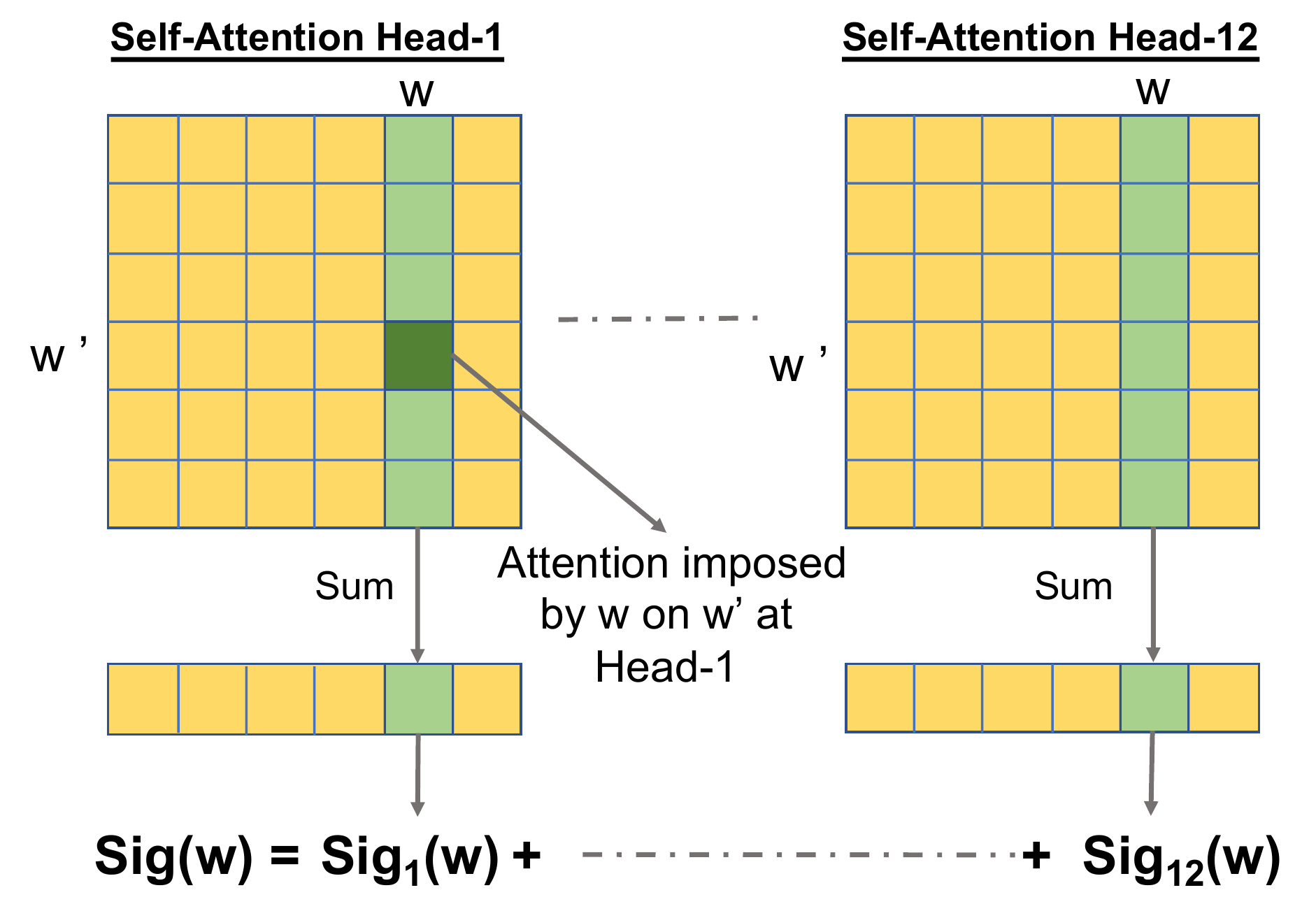}
}
\caption{Figure shows significance score computation for word-vector $w$ using the computed self-attention matrix.}
\label{fig:sig_score}
\end{figure}

\paragraph{Attention-based Scoring.}
Consider an encoder $j$ . 
In the {\pb} setting, the input matrix 
$\matM$ is of size $\ell_{j-1}\times 768$ and the attention 
matrices are of size $\ell_{j-1}\times \ell_{j-1}$.
Consider an attention head $h\in [1,12]$.
For a word $w'$, the row $\matZh[w',:]$ computed by the head $h$ can be written as $\sum_w \matAh[w', w] \cdot \matVh[w,:]$. 
In other words, the row $\matZh[w',:]$ is the weighted average of the rows of $\matVh$, taking the attention values as weights. 
Intuitively, we interpret the entry $\matAh[w', w]$ as the attention received by word $w'$ from $w$ on head $h$.

Our scoring function is based on the intuition that the significance of a word-vector $w$
can be estimated from the attention imposed by $w$ on the other word-vectors.
For a word-vector $w$ and a head $h$, we define the significance score
of $w$ for $h$ as $\sig_h(w) = \sum_{w'} \matAh[w', w]$.
The overall significance score of $w$ is then defined as the aggregate over the heads:
$\sig(w) = \sum_h \sig_h(w)$. Thus, the significance score
is the total amount of attention imposed by $w$ on the other words. 
See Figure \ref{fig:sig_score} for an illustration.

\begin{figure}[t]
\center
\resizebox{0.80\linewidth}{!}{
\includegraphics[width=2.5in]{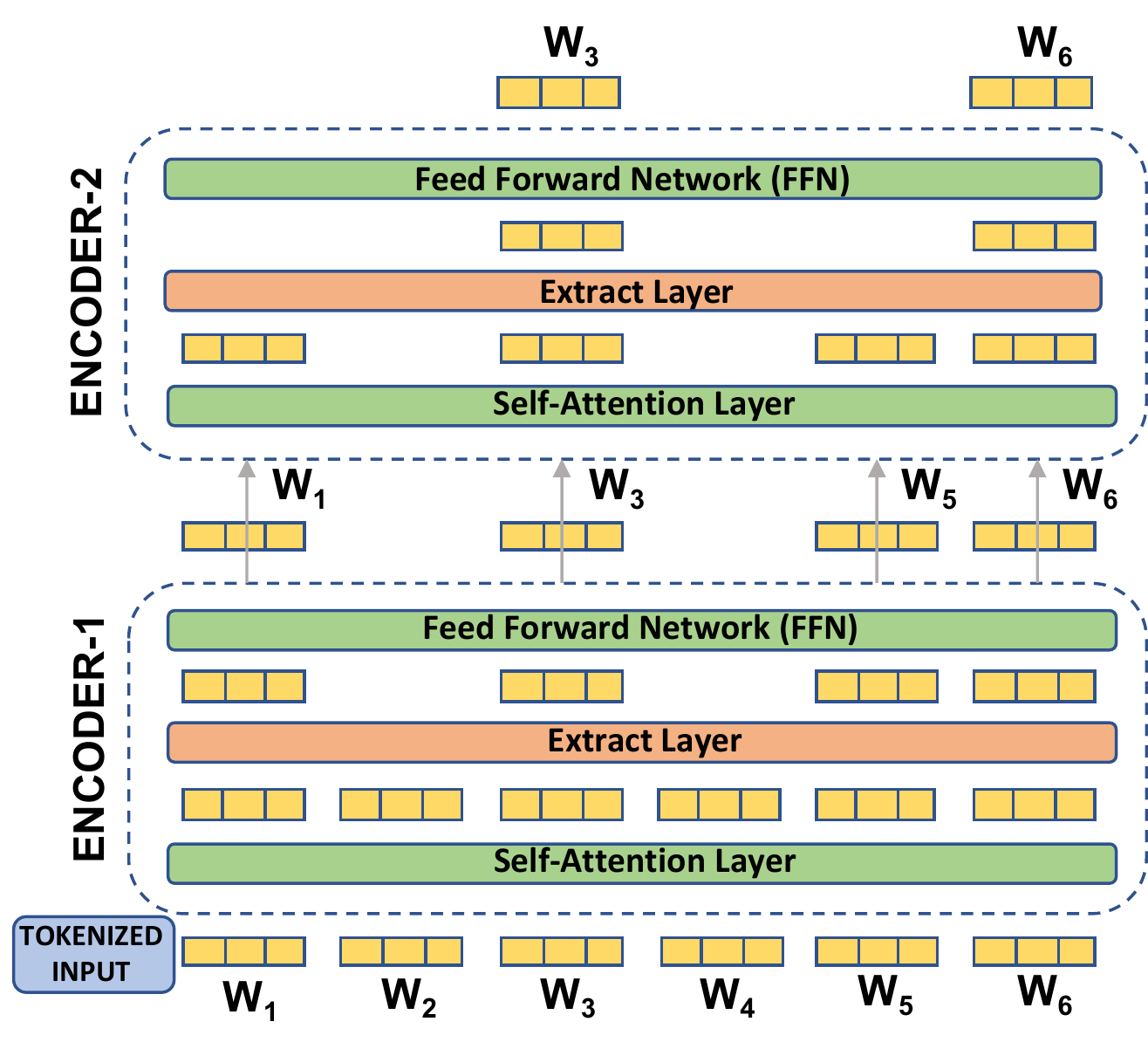}
}
\caption{
Word-vector selection over the first two encoders.
Here, $\maxl=6$, $\ell_1 = 4$ and $\ell_2 = 2$.  
The first encoder eliminates two word-vectors $w_2$ and $w_4$ with least significance scores; the second encoder further eliminates word-vectors $w_1$ and $w_5$. 
}
\label{fig:word_reduction}
\end{figure}

\paragraph{Word-vector Extraction.} 
Given the scoring mechanism, we perform word-vector selection 
by inserting an {\extract} layer between the self-attention module and the feed forward network.
The layer computes the scores and retains the top $\ell_j$ word-vectors.
See Figure \ref{fig:word_reduction} for an illustration.

\paragraph{Validation of the Scoring Function.} 
We conducted a study to validate the scoring function.
We utilized mutual information to analyze the effect of eliminating a single word-vector. 
The study showed that higher the score of the eliminated word-vector,
lower the agreement with the baseline model.
Thus, the scoring function satisfies the criterion we had aimed for:
the score of a word-vector is positively correlated with its influence on the final prediction. A detailed description of the study is deferred to the supplementary material.

{\pb} uses a scoring mechanism for estimating the significance of the word-vectors satisfying the following criterion: the score of a word-vector must be positively correlated with its influence on the final classification output (namely, word-vectors of higher influence get higher score). 

For a word-vector $w$ and a head $h$, we define the significance score
of $w$ for $h$ as $\sig_h(w) = \sum_{w'} \matAh[w', w]$ where $A_h$ is the {\em attention matrix} for head $h$ and $w'$ signifies other words in the input sequence.
The overall significance score of $w$ is then defined as the aggregate over the heads:
$\sig(w) = \sum_h \sig_h(w)$. 

We validate the scoring function using the well-known concept of mutual information
and show that the significance score of a word is positively correlated with the 
classification output.
Let $\bfX$ and $\bfY$ be two random variables. 
Recall that the {\em mutual information} between 
$\bfX$ and $\bfY$ is defined as $\MI(\bfX;\bfY) = \entropy(\bfX) - \entropy(\bfX|\bfY)$,
where $\entropy(\bfX)$ is the entropy of $\bfX$ and $\entropy(\bfX|\bfY)$ is the 
conditional entropy of $\bfX$ given $\bfY$. The quantity is symmetric with respect to the two variables.
Intuitively, Mutual information measure how much $\bfX$ and $\bfY$ agree with each other.
It quantifies the information that can be gained about $\bfX$ from information about $\bfY$.
If $\bfX$ and $\bfY$ are independent random variables, then $\MI(\bfX;\bfY) = 0$.
On the other hand, if the value of one variable can be determined with certainty given the
value of the other, then $\MI(\bfX;\bfY) = \entropy(\bfX) = \entropy(\bfY)$.

\begin{figure}[t!]
\center
\includegraphics[width=2.0in]{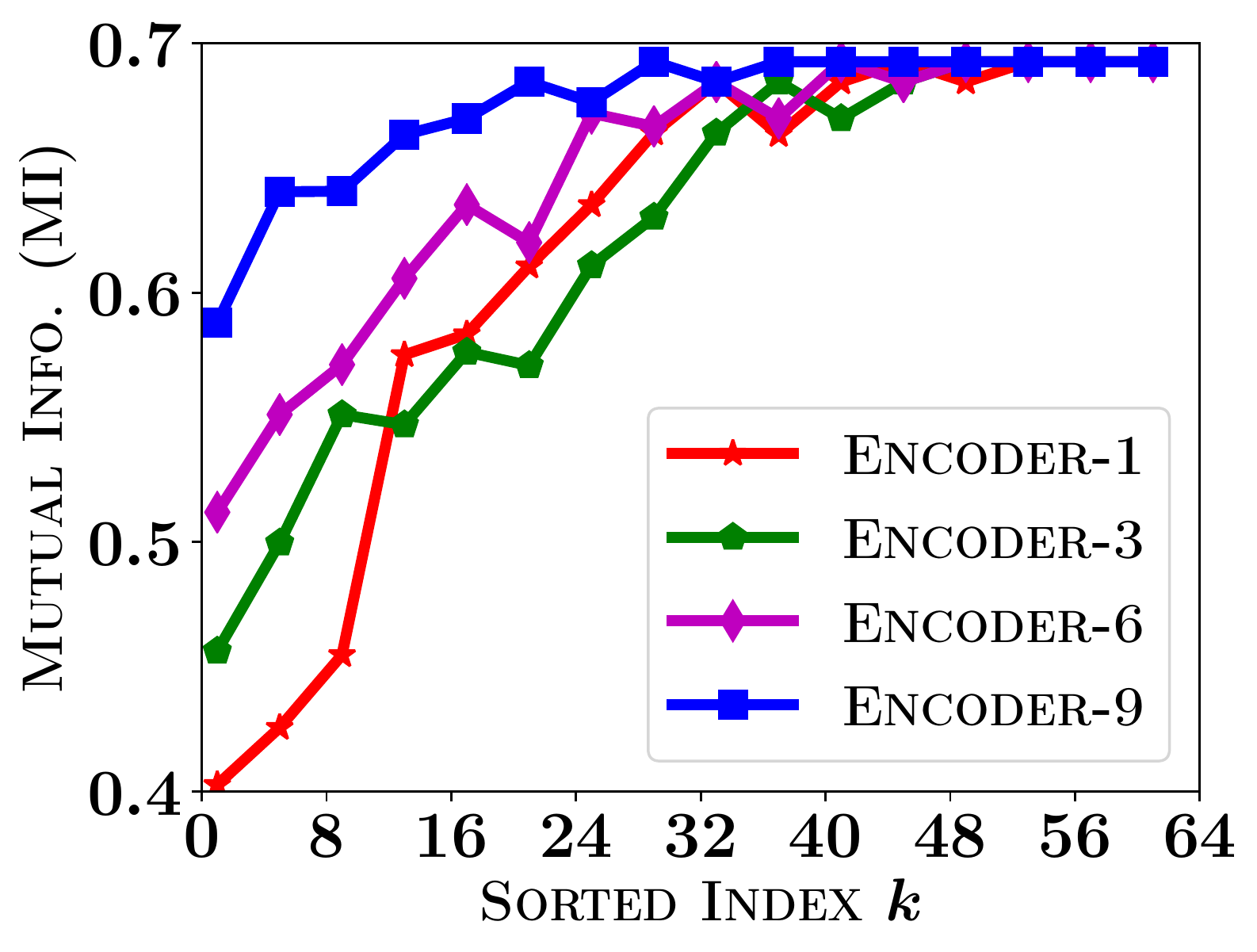}
\caption{Demonstration of mutual information}
\label{fig:mi}
\end{figure}

For this demonstration, we consider the SST-2 dataset from our experimental study with input length $\maxl=128$ 
and number of classification categories $C=2$ (binary classification).
Consider an encoder $j$ and we shall measure the mutual information between
the classification output of the original model and a modified model 
that eliminates a single word at encoder $j$.
Consider the trained {\bert} model without any word elimination and let $\bfX$ denote
the classification label output by the model on a randomly chosen input from the training data.
Fix an integer $k\leq \ell_{j-1}$ and let $w$ be the word with the $k^{th}$ highest significance score. 
Consider a modified model that does not eliminate any words on encoders $1, 2, \ldots, j-1$,
eliminates $w$ at encoder $j$, and does not eliminate any further words at encoders $j+1, j+2, \ldots, 12$.
Let $\bfY_k$ be the random variable that denotes the classification output of the above model on a randomly chosen input.
The mutual information between $\bfX$ and $\bfY_k$ can be computed using the formula:
\[
\sum_{b,b'\in \{0,1\}}
		\Pr(\bfX=b, \bfY=b')\cdot \ln\left[
		\frac
		{\Pr(\bfX=b, \bfY=b')}
		{\Pr(\bfX=b)\cdot \Pr(\bfY=b')}\right]
\]

We measured the mutual information between $\bfX$ and $\bfY_k$ for all encoders $j$ and for all $k\in [1,128]$.
Figure \ref{fig:mi} shows the above data, wherein for the simplicity of presentation, we have restricted to encoders $j=1,3,6,9$.
In this case, since the model predictions are approximately balanced between positive and negative, the baseline entropy is $\entropy(\bfX) \sim \ln(2) = 0.69$.
We can make two observations from the figure. 
First is that as $k$ increases, the mutual information increases. 
This implies that deleting words with higher score results in higher loss of mutual information. 
Alternatively, deleting words with lower score results in higher mutual information, 
meaning the modified model gets in better agreement with the original model.
Secondly, as the encoder number $j$ increases, the mutual information approaches the baseline entropy faster,
confirming our hypothesis that words of higher significance scores (or more words) can be eliminated from the later encoders.
The figure demonstrates that the score $\sig(\cdot)$ captures the significance of the words in an effective manner.

\subsection{Retention Configuration}
\label{sec:config}
We next address the task of determining the retention configuration.
Analyzing all the possible configurations is untenable due to the exponential search space.
Instead, we design a strategy that learns the retention configuration.
Intuitively, we wish to retain the word-vectors with the topmost significance
scores and the objective is to learn how many to retain.
The topmost word-vectors may appear in arbitrary positions
across different inputs in the dataset.
Therefore, we sort them according to their significance scores.
We shall learn the extent to which the sorted positions must be retained.
We accomplish the task by introducing {\softextract} layers 
and modifying the loss function.

\paragraph{Soft-extract Layer.}
The {\extract} layer either selects or eliminates a word-vector (based on scores).
In contrast, the {\softextract} layer would retain all the word-vectors, 
but to varying degrees as determined by their significance.

Consider an encoder $j$ and let $w_1, w_2, \ldots, w_{\maxl}$ be the
sequence of word-vectors input to the encoder. The significance
score of $w_i$ is given by $\sig(w_i)$. 
Sort the word-vectors in the decreasing order of their scores.
For a word-vector $w_i$, let $\sigidx(w_i)$ denote the position of $w_i$ in the sorted order;
we refer to it as the {\em sorted position} of $w_i$.

\begin{figure}[t]
\center
\resizebox{0.9\linewidth}{!}{
\includegraphics[width=2.5in]{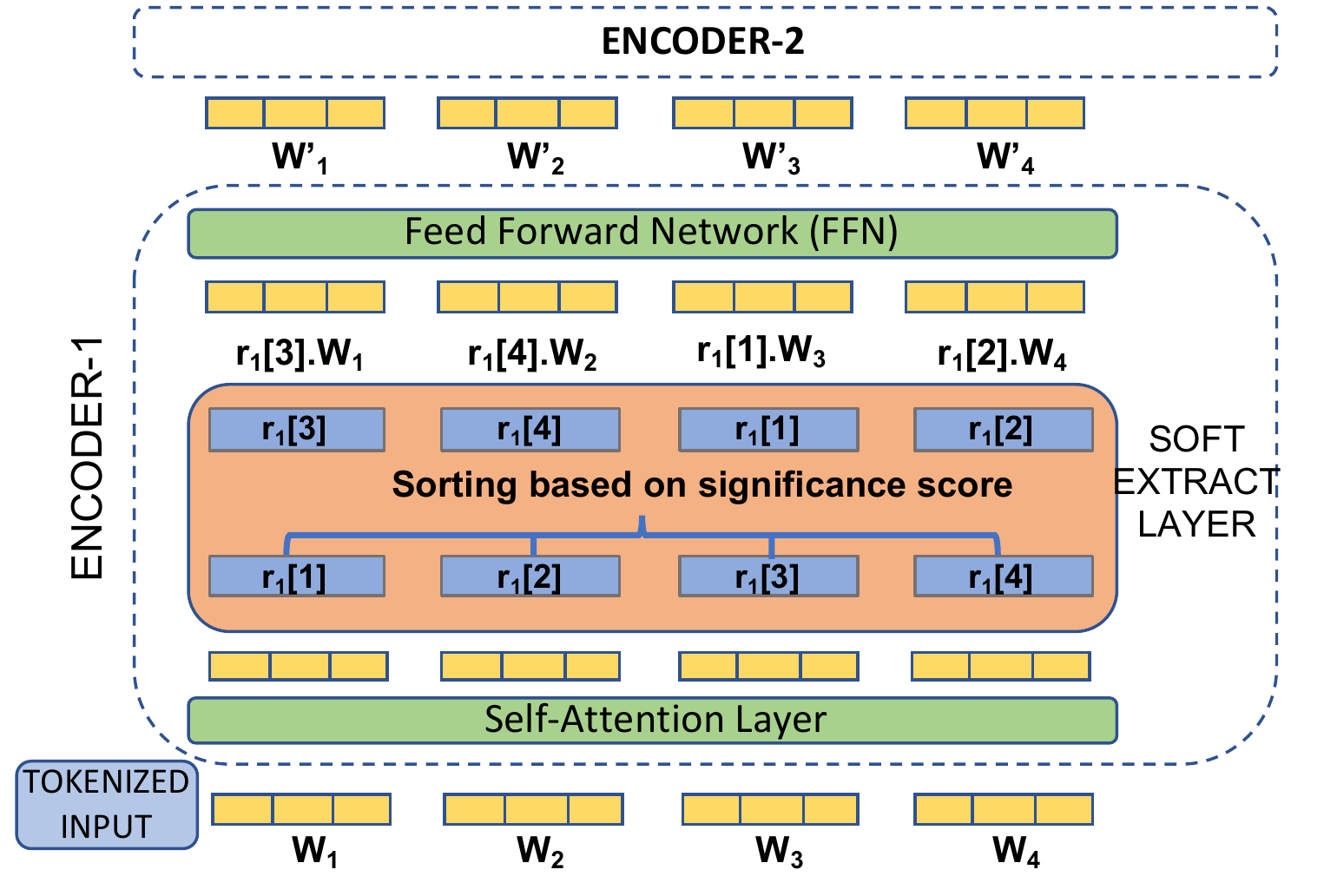}
}
\caption{{\softextract} layer. First encoder is shown, taking $\maxl=4$.
In this example, the sorted sequence of the word-vectors is $w_3, w_4, w_1, w_2$;
the most significant word-vector $w_3$ gets multiplied by $\sigparam_1[1]$
and the least significant word-vector $w_2$ by $\sigparam_1[4]$.
}
\label{fig:soft-extract}
\end{figure}

The {\softextract} layer involves $\maxl$ learnable parameters,
denoted $\sigparam_j[1], \ldots, \sigparam_j[\maxl]$, called retention parameters.
The parameters are constrained to be in the range $[0,1]$.
Intuitively, the parameter $\sigparam_j[k]$ represents 
the {\em extent} to which the $k^{th}$ sorted position is retained.

The {\softextract} layer is added in between the self-attention
module and the feed forward network, and performs the following transformation.
Let $\matE^{in}$ denote the matrix of size $\maxl\times 768$ output by the self-attention layer. For $i\in [1,\maxl]$, the row $\matE^{in}[i,:]$ yields the word-vector $w_i$. 
The layer multiplies the word-vector by the retention parameter corresponding
to its sorted position:
\[
	\matE^{out}[i,:] = \sigparam_j[\sigidx(w_i)] \cdot \matE^{in}[i,:].
\]
The modified matrix $\matE^{out}[i,:]$ is input to the feed-forward network.
The transformation ensures that all the word-vectors in the $k^{th}$ sorted
position get multiplied by the same parameter $\sigparam_j[k]$.
Figure \ref{fig:soft-extract} presents an illustration.

\paragraph{Loss Function.}
We define the {\em mass} at encoder $j$ to be the extent to which the sorted positions
are retained, i.e., ${\mass}(j;\sigparam) = \sum_{k=1}^{\maxl} \sigparam_j[k]$.
Our aim is to minimize the aggregate mass over all the encoders with minimal
loss in accuracy.
Intuitively, the aggregate mass may be viewed as a budget on the total number of positions retained;
$\mass(j;\sigparam)$ is the breakup across the encoders.

We modify the loss function by incorporating
an $L_1$ regularizer over the aggregate mass.
As demonstrated earlier, the encoders have varying influence on the classification output. We scale the 
mass of each encoder by its index.
Let $\mytheta$ denote the parameters of the baseline {\bert} model and $\calL(\cdot)$
be the loss function (such as cross entropy loss or mean-squared error)
as defined in the original task. We define the new objective function as:
\begin{eqnarray*}
&&\min_{\mytheta, \sigparam} 
\left[
{\calL}(\mytheta, \sigparam)~+ ~\lambda \cdot \sum_{j=1}^{L} j\cdot \mass(j;\sigparam)
\right]\\
&& {\rm s.t.}~~\sigparam_j[k]\in [0,1]~~\forall(j\in [1,L], k\in [1,\maxl]),
\end{eqnarray*}
where $L$ is the number of encoders.
While $\calL(\mytheta,\sigparam)$ controls the accuracy, 
the regularizer term controls the aggregate mass. The hyper-parameter $\lambda$ tunes the trade-off.
%
%

The retention parameters are initialized as $\sigparam_j[k] = 1$, meaning 
all the sorted positions are fully retained to start with.
We train the model to learn the retention parameters.
The learned parameter $\sigparam_j[k]$ provides
the extent to which the word-vectors at the $k^{th}$ sorted position must be retained.
We obtain the retention configuration from the mass of the above parameters:
for each encoder $j$, set $\ell_j = {\tt ceil}(\mass(j))$.
In the rare case where the configuration is non-monotonic, 
we assign $\ell_j = \min\{\ell_j, \ell_{j-1}\}$.

\subsection{Training PoWER-BERT}
\label{sec:training}
Given a dataset, the scheme involves three training steps:
\begin{enumerate}
\item {\it Fine-tuning:} Start with the pre-trained {\bert} model and fine-tune it on the given dataset.
\item {\it Configuration-search:} Construct an auxiliary model by inserting the {\softextract} layers
in the fine tuned model, and modifying its loss function. 
The regularizer parameter $\lambda$ is tuned to derive
the desired trade-off between accuracy and inference time. 
The model consists of parameters of the original {\bert} model
and the newly introduced {\softextract} layer.
We use a higher learning rate for the latter.
We train the model and derive the retention configuration.
\item {\it Re-training:} Substitute the {\softextract} layer by {\extract} layers.
The number of word-vectors to retain at each encoder is determined
by the retention configuration computed in the previous step.
The word-vectors to be retained are selected based on their significance scores.
We re-train the model. 
\end{enumerate}
In our experiments, all the three steps required only $2-3$ epochs.
Inference is performed using the re-trained {\pb} model. 
The {\cls} token is never eliminated and it is used to derive the final prediction.

\begin{table}[b!]
\caption{Dataset statistics: NLI and QA refers to Natural Language Inference and Question Answering tasks respectively. Note that STS-B is a regression task, therefore doesn't have classes.}
\begin{center}
\begin{small}
\begin{sc}
\resizebox{0.9\linewidth}{!}{
\begin{tabular}{lccc}
\toprule
\multirow{2}{*}{Dataset} & \multirow{2}{*}{Task} & \multirow{2}{*}{$\#$ Classes} & Input Seq. \\
&  &  & Length $(N)$\\
\midrule
CoLA       & Acceptability  & 2   & 64 \\ 
RTE        & NLI            & 2   & 256 \\  
QQP        & Similarity     & 2   & 128 \\
MRPC       & Paraphrase     & 2   & 128 \\ 
SST-2      & Sentiment      & 2   & 64 \\ 
MNLI-m     & NLI            & 3   & 128 \\ 
MNLI-mm    & NLI            & 3   & 128 \\ 
QNLI       & QA/NLI         & 2   & 128 \\   
STS-B      & Similarity     & -   & 64 \\  
\midrule
IMDB       & Sentiment      & 2   & 512 \\ 
RACE       & QA             & 2   & 512 \\ 
\bottomrule
\end{tabular}
}
\end{sc}
\end{small}
\end{center}
\label{tab:dataset_statistics}
\end{table}

\begin{table*}[t]
\caption{Comparison between {\pb} and {\bertbase}. We limit the accuracy loss for {\pb} to be within $1\%$ by tuning the regularizer parameter $\lambda$. Inference done on a K80 GPU with batch size of 128 (averaged over 100 runs). Matthew's Correlation reported for CoLA; F1-score for QQP and MRPC; Spearman Correlation for STS-B; Accuracy for the rest.}
\begin{center}
\begin{large}
\begin{sc}
\resizebox{\linewidth}{!}{
\renewcommand{\arraystretch}{1.2}
\begin{tabular}{l|l|ccccccccccc}
\toprule
 & Method & CoLA  & RTE & QQP & MRPC & SST-2 & MNLI-m & MNLI-mm & QNLI & STS-B & IMDB & RACE \\
 \midrule
\multirow{2}{*}{Test Accuracy} & \multicolumn{1}{l|}{\bertbase} & 52.5 & 68.1 & 71.2 & 88.7 & 93.0 & 84.6 & 84.0 & 91.0 & 85.8  & 93.5 & 66.9 \\
 & \multicolumn{1}{l|}{{\pb}} & 52.3 & 67.4 & 70.2 & 88.1 & 92.1 & 83.8 & 83.1 & 90.1 & 85.1 & 92.5 & 66.0 \\ 
 
\midrule
\multirow{2}{*}{Inference Time (ms)} & \multicolumn{1}{l|}{\bertbase} & 898 & 3993 & 1833 & 1798 & 905 & 1867 & 1881 & 1848 & 881 & 9110 & 20040 \\ 
& \multicolumn{1}{l|}{{\pb}} & 201 & 1189 & 405 & 674 & 374 & 725 & 908 & 916 & 448 & 3419 & 10110 \\ 

\midrule
Speedup & & \textbf{(4.5x)} & (3.4x) & \textbf{(4.5x)} & (2.7x) & (2.4x) & (2.6x) & (2.1x) & (2.0x) & (2.0x) & (2.7x) & (2.0x) \\ 
\bottomrule
\end{tabular}
}
\end{sc}
\end{large}
\end{center}
\label{tab:BERT_PoWER-BERT}
\end{table*}

\begin{table*}[t]
\caption{Comparison between {\pb} and {\albert}. Here {\pb} represents application of our scheme on {\albert}. The experimental setup is same as in Table \ref{tab:BERT_PoWER-BERT}}
\begin{center}
\begin{large}
\begin{sc}
\resizebox{0.9\linewidth}{!}{
\renewcommand{\arraystretch}{1.2}
\begin{tabular}{l|c|ccccccccc}
\toprule
 & Method & CoLA  & RTE & QQP & MRPC & SST-2 & MNLI-m & MNLI-mm & QNLI & STS-B\\
 
 \midrule
\multirow{2}{*}{Test Accuracy} & \multicolumn{1}{c|}{{\albert}} & 42.8 & 65.6 & 68.3 & 89.0 & 93.7 & 82.6 & 82.5 & 89.2 & 80.9 \\ 
 & \multicolumn{1}{c|}{{\pb}} & 43.8 & 64.6 & 67.4 & 88.1 & 92.7  & 81.8 & 81.6 & 89.1 & 80.0\\ 
 
\midrule
\multirow{2}{*}{Inference Time (ms)} & \multicolumn{1}{c|}{{\albert}} & 940 & 4210 & 1950 & 1957 & 922 & 1960 & 1981 & 1964 & 956 \\ 
 & \multicolumn{1}{c|}{{\pb}} & 165 & 1778 & 287 & 813 & 442 & 589 & 922 & 1049& 604\\ 
 \midrule
Speedup &  & (5.7x) & (2.4x) & \textbf{(6.8x)} & (2.4x) & (2.1x) & (3.3x) & (2.1x) & (1.9x) & (1.6x) \\
\bottomrule
\end{tabular}
}
\end{sc}
\end{large}
\end{center}
\label{tab:ALBERT_PoWER-BERT}
\end{table*}

\section{Experimental Evaluation}
\subsection{Setup}
\paragraph{Datasets.}
We evaluate our approach on a wide spectrum of classification/regression tasks
pertaining to $9$ datasets from the GLUE benchmark \cite{glue},
and the IMDB \cite{maas-EtAlACL-HLT2011} and the RACE \cite{lai2017large}) datasets.
The datasets details are shown in Table \ref{tab:dataset_statistics}. 

\paragraph{Baseline methods.}
We compare {\pb} with the state-of-the-art inference time reduction methods: 
{\distilbert} \cite{distil-bert}, {\bertpkd} \cite{bert-pkd} and {\headprune} \cite{head-prune}. 
They operate by removing the parameters:
the first two eliminate encoders, and the last prunes attention heads.
Publicly available implementations were used for these methods \cite{DistilBERT-Code, BERT-PKD-Code, Head-Prune-Code}. 

\paragraph{Hyper-parameters and Evaluation.}
Training {\pb} primarily involves four hyper-parameters, which we select from the ranges listed below:
a) learning rate for the newly introduced {\softextract} layers - $[10^{-4}, 10^{-2}]$; 
b) learning rate for the parameters from the original {\bert} model - $[2\times 10^{-5}, 6\times 10^{-5}]$;
c) regularization parameter $\lambda$ that controls the  trade-off  between  accuracy  and  inference  time - $[10^{-4}, 10^{-3}]$; 
d) batch size - $\{4, 8, 16, 32, 64\}$. 
Hyper-parameters specific to the datasets are provided in the supplementary material.

The hyper-parameters for both {\pb} and the baseline methods were tuned on the Dev dataset for 
GLUE and RACE tasks. For IMDB, we subdivided the training data into 80\% for training and 20\% for tuning.
The test accuracy results for the GLUE datasets were obtained by submitting the predictions to the evaluation server \footnote{\url{https://gluebenchmark.com}},
whereas for IMDB and RACE, the reported results are on the publicly available Test data.

\paragraph{Implementation.} The code for {\pb} was implemented in Keras and is available at \url{https://github.com/IBM/PoWER-BERT}. 
The inference time experiments for {\pb} and the baselines were conducted using Keras framework on a K80 GPU machine. A batch size of 128 (averaged over 100 runs)
was used for all the datasets except RACE, for which the batch size was set to 32 (since each input question has 4 choices of answers).

\paragraph{Maximum Input Sequence Length.}
\label{sec:ell}
The input sequences  are of varying length and are padded to get a uniform length of $\maxl$.
Prior work use different values of $\maxl$,
for instance {\albert} uses $\maxl=512$ for all GLUE datasets.
However, only a small fraction of the inputs are of length close to the maximum.
Large values of $\maxl$ would offer easy pruning opportunities and larger gains for {\pb}.
To make the baselines competitive, we set stringent values of $\maxl$: 
we determined the length $\maxl'$ such that at most $1\%$ of the input sequences are longer than $\maxl'$ and fixed $\maxl$ to be the value from $\{64, 128, 256, 512\}$
closest to $\maxl'$. Table \ref{tab:dataset_statistics} presents the lengths specific to each dataset.

\begin{figure*}[t]
\centering
\resizebox{0.9\linewidth}{!}{
\begin{tabular}{ccc}
\includegraphics[width=2.0in]{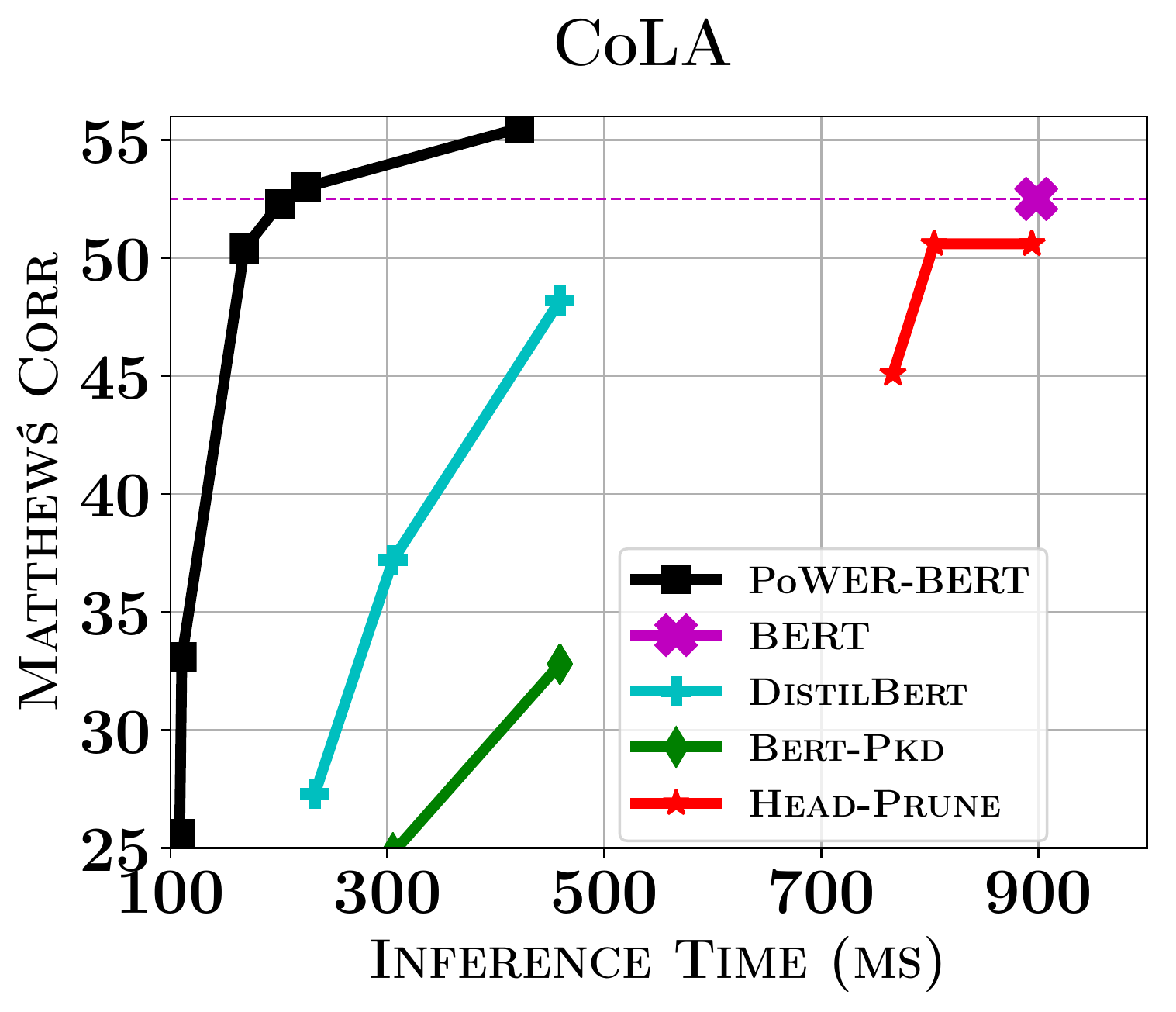}
&
\includegraphics[width=2.0in]{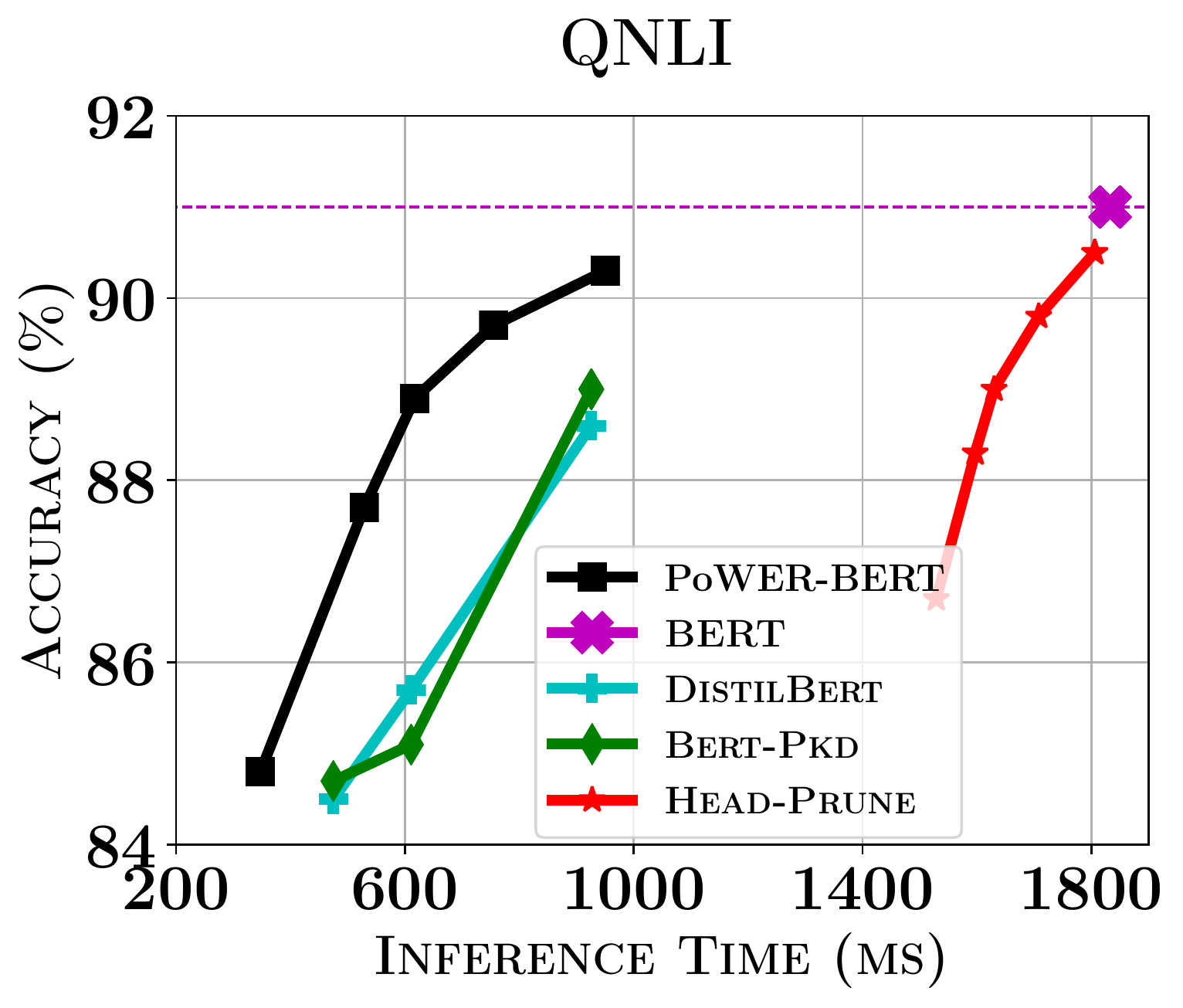}
&
\includegraphics[width=2.0in]{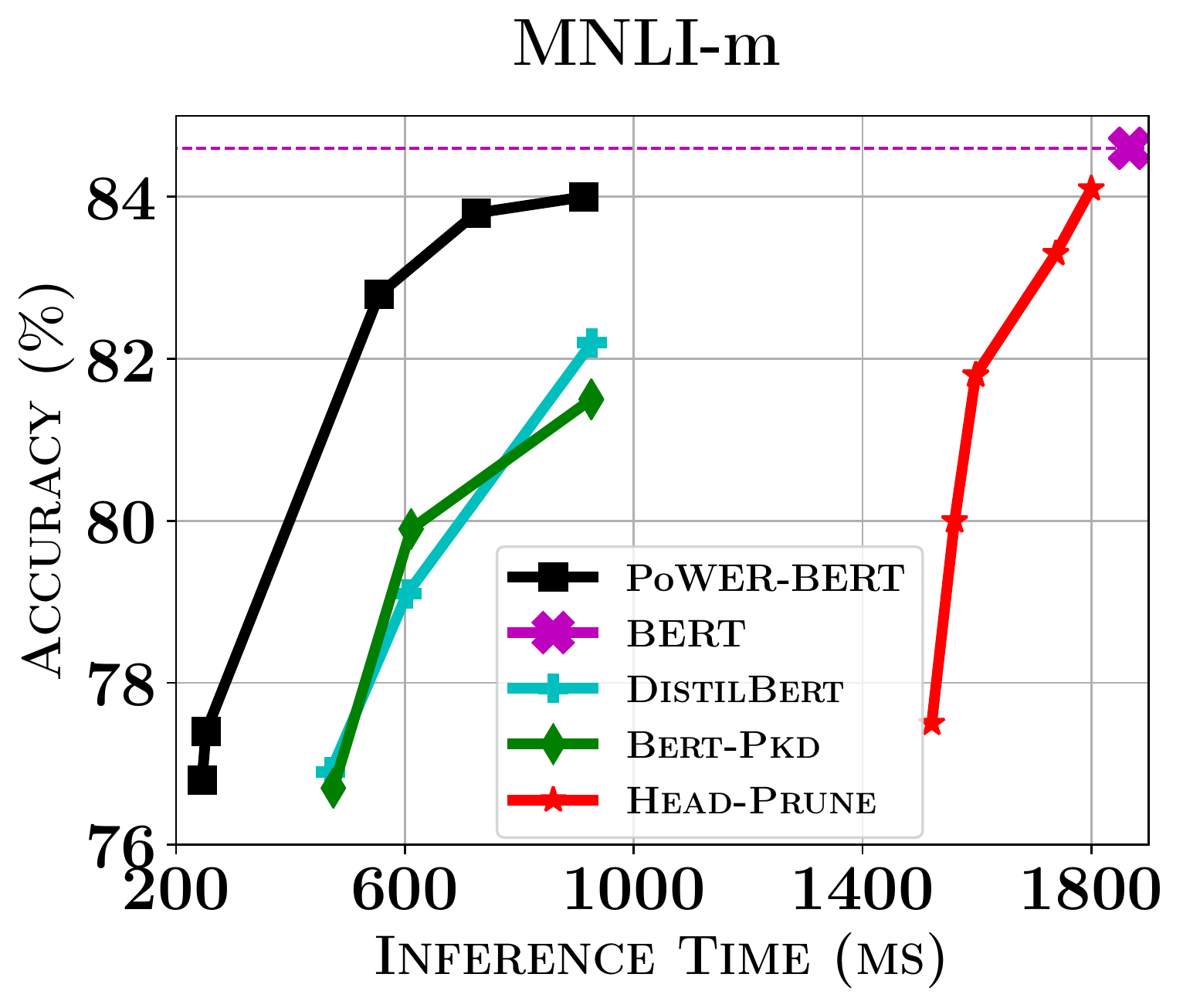}
\\
\includegraphics[width=2.0in]{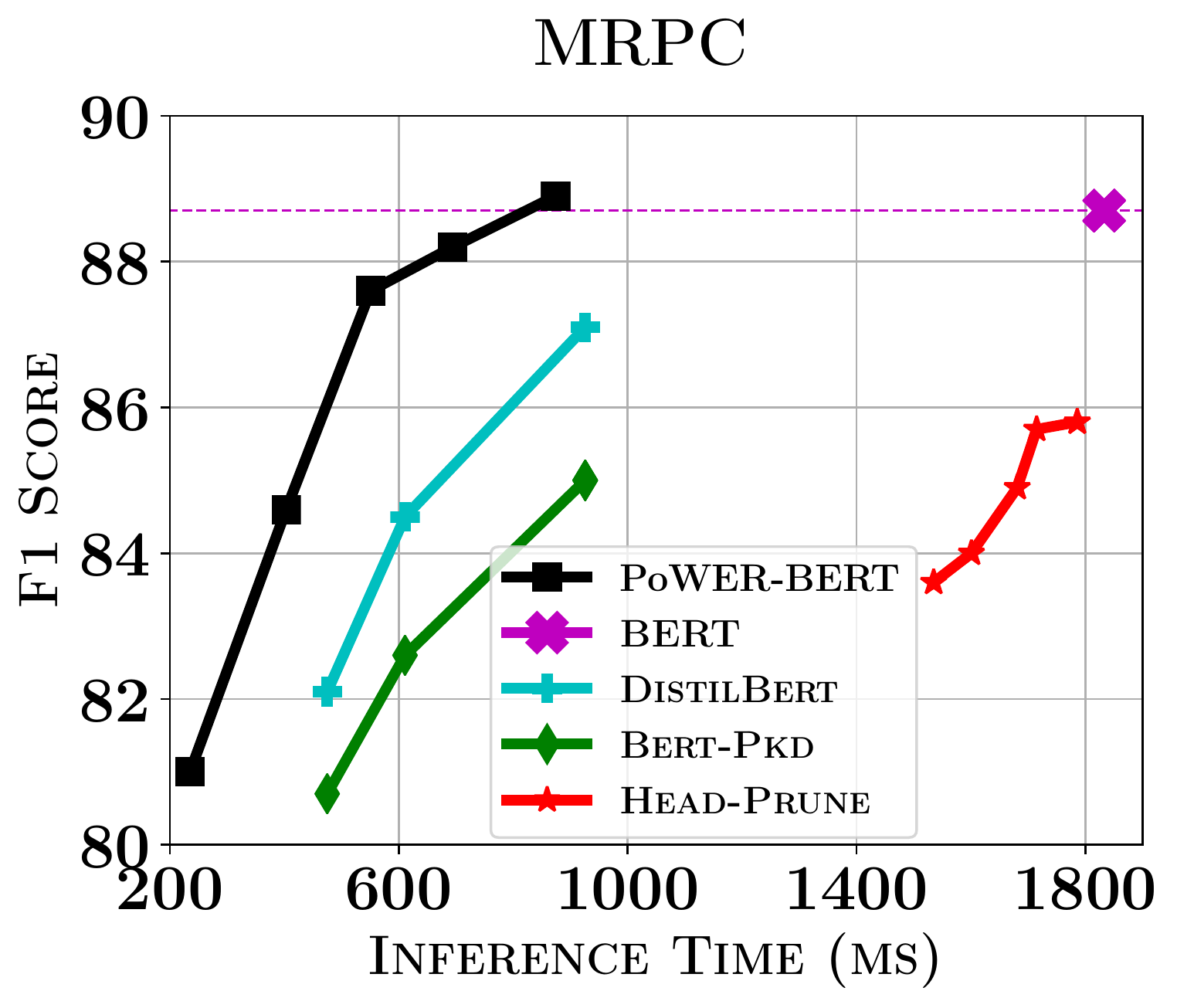}
&
\includegraphics[width=2.0in]{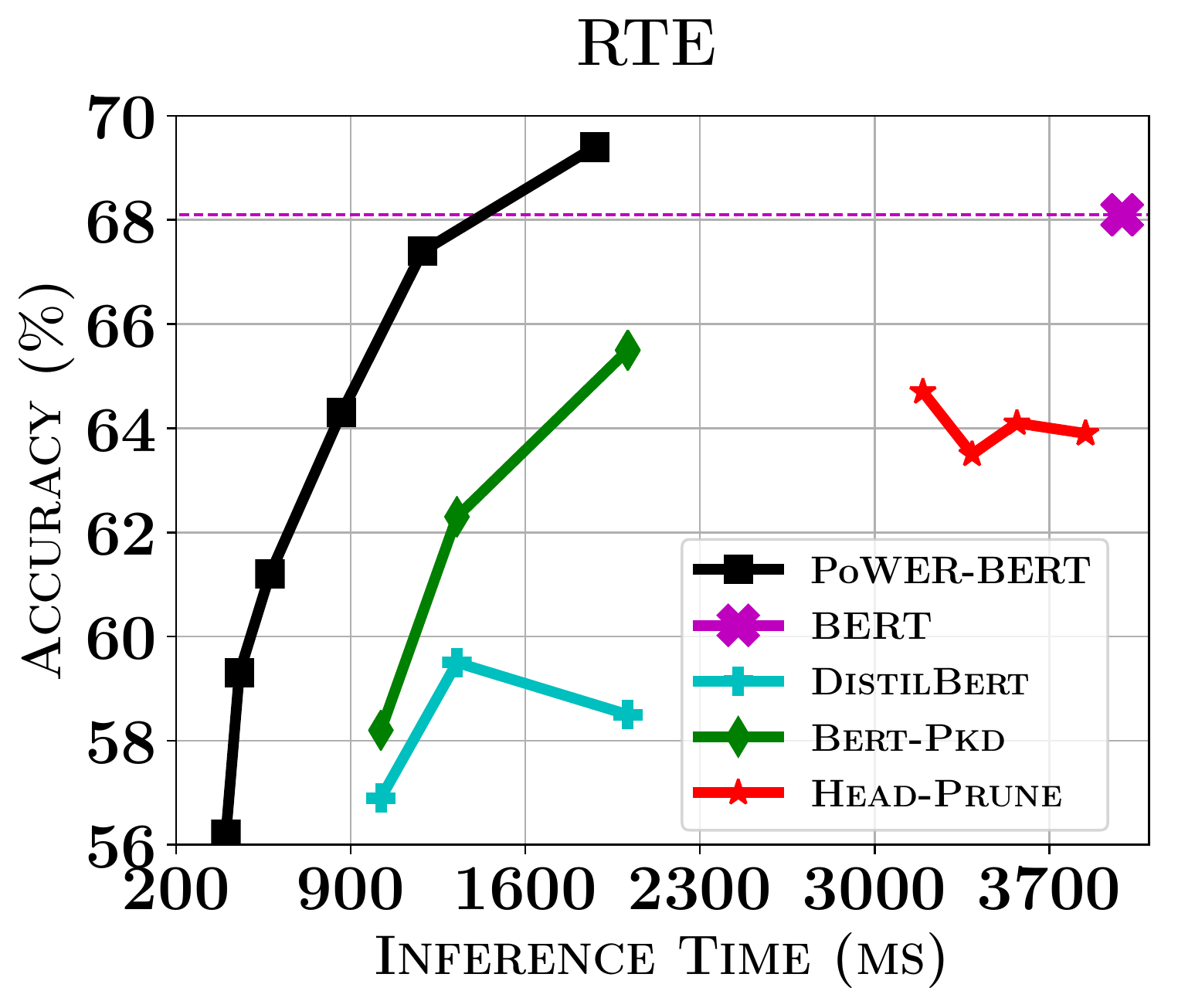}
&
\includegraphics[height=1.75in, width=2.0in]{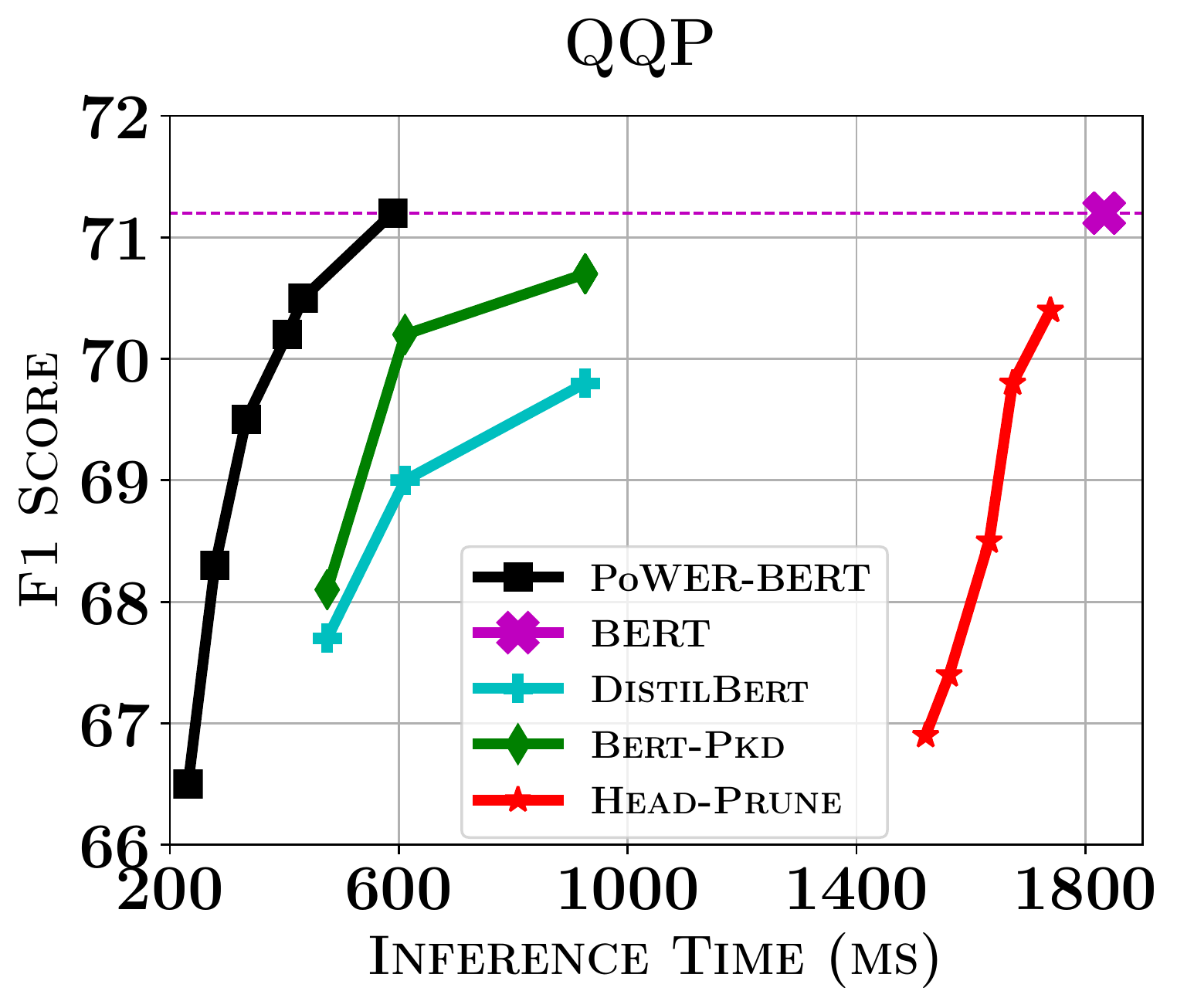}
\end{tabular}
}
\caption{
Comparison to prior methods. Pareto curves showing accuracy vs. inference time 
trade-off. Top-left corners correspond to the best inference time and accuracy.
Points for {\pb} obtained by tuning the regularizer parameter $\lambda$.
For {\distilbert} and {\bertpkd}, the points correspond to retaining $\{3,4,6\}$
encoders. For {\headprune}, points obtained by varying number of retained attention-heads.
The cross represents {\bertbase} performance;
dotted line represents its accuracy (for the ease of comparison). 
Over the best baseline method, {\pb} offers:
accuracy gains as high as $16\%$ on CoLA and $6\%$ on RTE at inference time $305$ ms 
and $1326$ ms, respectively;
inference time gains as high as $2.7$x on CoLA and $2$x on RTE at accuracy $48.2\%$
and 
$65.5\%$, respectively.
} 
\label{fig:8chart}
\end{figure*}
\subsection{Evaluations}
\paragraph{Comparison to BERT.}
In the first experiment, we demonstrate the effectiveness of the word-vector elimination
approach by evaluating the inference time gains achieved by {\pb} over {\bertbase}.
We limit the accuracy loss to be within $1\%$ by tuning the regularizer parameter $\lambda$
that controls the trade-off between inference time and accuracy.
The results are shown in Table \ref{tab:BERT_PoWER-BERT}.
We observe that {\pb} offers at least $2.0$x reduction in 
inference time on all the datasets and the improvement can be as high as $4.5$x,
as exhibited on the CoLA and the QQP datasets.

We present an illustrative analysis by considering the RTE dataset.
The input sequence length for the dataset is $\maxl=256$.
Hence, across the twelve encoders, {\bertbase} needs to process $12\times 256 = 3072$ word-vectors for any input.
In contrast, the retention configuration used by {\pb} on this dataset
happens to be $(153, 125, 111, 105, 85, 80, 72, 48, 35, 27, 22, 5)$ summing to $868$.
Thus, {\pb}  processes an aggregate of only $868$ word-vectors.
The self-attention and the feed forward network modules of the encoders
perform a fixed amount of computations for each word-vector.
Consequently, the elimination of the word-vectors
leads to reduction in computational load and improved inference time.

\paragraph{Comparison to Prior Methods.} 
In the next experiment, we
compare {\pb} with the state-of-the-art inference time reduction methods, 
by studying the trade-off between accuracy and inference time. 
The Pareto curves for six of the GLUE datasets are shown in Figure \ref{fig:8chart};
others are provided in the supplementary material.
Top-left corners correspond to the best inference time and accuracy.

For {\pb}, the points on the curves were obtained by tuning the regularizer parameter $\lambda$. For the two encoder elimination methods, {\distilbert} and {\bertpkd},
we derived three points by retaining $3$, $4$, and $6$ encoders;
these choices were made so as to achieve inference time gains comparable to {\pb}.
Similarly, for the {\headprune} strategy, the points were obtained by varying 
the number of attention-heads retained.

Figure \ref{fig:8chart} demonstrates that {\pb} exhibits marked dominance over 
all the prior methods offering:
\begin{itemize}
\item
Accuracy gains as high as $16\%$ on CoLA and $6\%$ on RTE for a given inference time.
\item
Inference time gains as high as $2.7$x on CoLA and $2.0$x on RTE for a given accuracy.
\end{itemize}

The results validate our hypothesis
that fine-grained word-vector elimination yields better trade-off than
coarse-grained encoder elimination. We also observe that 
{\headprune} is not competitive. The reason is 
that the method exclusively targets the attention-heads
constituting only $~26\%$ of the {\bertbase} parameters
and furthermore, pruning a large fraction of the heads 
would obliterate the critical self-attention mechanism of {\bert}.

\paragraph{Accelerating ALBERT.} 
As discussed earlier, 
word-vector elimination scheme can be applied over compressed models as well.
To demonstrate, we apply {\pb} over {\albert},
one of the best known compression methods for {\bert}.
The results are shown in Table \ref{tab:ALBERT_PoWER-BERT} for the GLUE datasets.
We observe that the {\pb} strategy is able to accelerate
{\albert} inference by $2$x factors on most of the datasets (with $<1\%$ loss in accuracy), 
with the gain being as high as $6.8$x on the QQP dataset.

\begin{table}[]
\caption{Comparison of the accuracy of the word-vector selection methods on the SST-2 Dev set for a fixed retention configuration.}
\begin{center}
\begin{small}
\begin{sc}
\resizebox{\linewidth}{!}{
\renewcommand{\arraystretch}{1.2}
\begin{tabular}{lcccc}
\toprule
 & {\headws} & {\randws} & {\attnws}\\
\midrule
Entire dataset & 85.4\% & 85.7\% & 88.3\% \\
Input sequence length $> 16$ & 83.7\% & 83.4\% & 87.4\% \\
\bottomrule
\end{tabular}
}
\end{sc}
\end{small}
\end{center}
\label{tab:Word-selection}
\end{table}

\paragraph{Ablation Study.}
In Section \ref{sec:wsel}, we described three methods for word-vector selection:
two static techniques, {\headws} and {\randws},
and a dynamic strategy, denoted {\attnws}, based on the significance
scores derived from the attention mechanism.
We demonstrate the advantage of
{\attnws} by taking the SST-2 dataset as an illustrative example.
For all the three methods, we used the same sample retention configuration of 
$(64, 32, 16, 16, 16, 16, 16, 16, 16)$. 
The accuracy results are shown in Table \ref{tab:Word-selection}.
The first row of the table shows that {\attnws} offers improved accuracy.
We perform a deeper analysis by filtering inputs based on length. In the given sample configuration, most encoders retain only $16$ word-vectors. Consequently, we selected a threshold of $16$ and considered a restricted dataset with inputs longer than the threshold. The second row shows the accuracy results.

Recall that  {\headws}
relies on eliminating as many {\pad} tokens as possible on the average.
We find that the strategy fails on longer inputs, since many important word-vectors may get eliminated.
Similarly, {\randws} also performs poorly, since it is oblivious to the importance of the word-vectors. In contrast, {\attnws} achieves higher accuracy by carefully selecting word-vectors based on their significance. The inference time is the same for all the methods, as the same number of word-vectors get eliminated.

\begin{figure}[t]
\centering
\resizebox{\linewidth}{!}{
\includegraphics[width=2.0in]{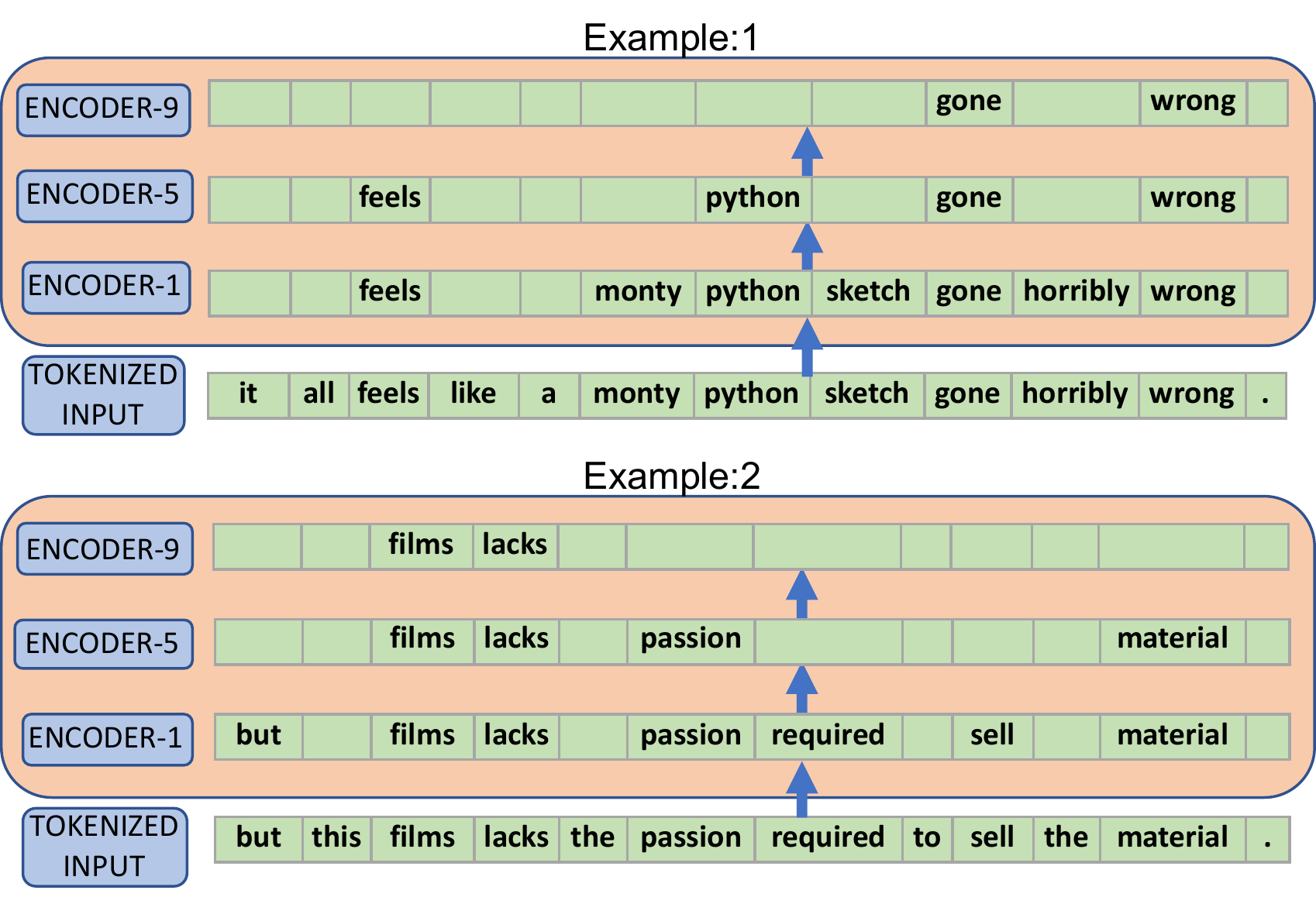}
}
\caption{Anecdotal Examples. Real-life examples from SST-2 dataset demonstrating progressive word-vector elimination.} 
\label{fig:anec}
\end{figure}

\paragraph{Anecdotal Examples. }
We present real-life examples 
demonstrating word-vector redundancy 
and our word-vector selection strategy based on the self-attention mechanism (\attnws).
For this purpose, we experimented with sentences 
from the SST-2 sentiment classification dataset
and the results are shown in Figure \ref{fig:anec}.

Both the sentences have input sequence length $\maxl=12$ (tokens). 
We set the retention configuration
as $(7,7,7,7,4,4,4,4,2,2,2,2)$
so that it progressively removes 
five word-vectors
at the first encoder, and two more at the fifth and the ninth encoders, each.

In both the examples, the first encoder eliminates the word-vectors
corresponding to stop words and punctuation. 
The later encoders may seem to eliminate more relevant word-vectors.
However, their information is captured by the word-vectors
retained at the final encoder, due to the diffusion of information.
These retained word-vectors carry the sentiment
of the sentence  and are sufficient for correct prediction.
The above study further reinforces our premise
that word-vector redundancy can be exploited
to improve inference time, while maintaining accuracy.

\section{Conclusions}
We presented {\pb}, a novel method for improving the inference time of the {\bert} model by exploiting word-vectors redundancy. Experiments on the standard GLUE benchmark show that {\pb} achieves up to $4.5$x gain in inference time over {\bertbase} with $<1\%$ loss in accuracy. Compared to prior techniques, it offers significantly better trade-off between accuracy and inference time. We showed that our scheme can be applied over {\albert}, a highly compressed variant of {\bert}. For future work, we plan to extend {\pb} to wider range of tasks such as language translation and text summarization.

\newpage
\bibliographystyle{named}
\bibliography{main}

\end{document}